\pdfoutput=1

\documentclass[11pt]{article}

\usepackage[final]{emnlp2021}
\usepackage{outlines}

\usepackage{times}
\usepackage{latexsym}
\usepackage{graphicx}
\usepackage{float}
\usepackage[T1]{fontenc}

\usepackage[utf8]{inputenc}

\usepackage{microtype}

\title{Machine Translation Impact in E-commerce Multilingual Search }

\author{Bryan Hang Zhang \\
  Amazon.com \\
  \texttt{bryzhang@amazon.com} \\\And
  Amita Misra \\
  Amazon.com\\
  \texttt{misrami@amazon.com} \\
}

\begin{document}
\maketitle
\begin{abstract}
 Previous work suggests that performance of cross-lingual information retrieval correlates highly with the quality of Machine Translation. However, there may be a threshold beyond which improving query translation quality yields little or no benefit to further improve the retrieval performance. This threshold may depend upon multiple factors including the source and target languages, the existing MT system quality and the search pipeline. In order to identify the benefit of improving an MT system for a given search pipeline, we investigate the sensitivity of retrieval quality to the presence of different levels of MT quality using experimental datasets collected from actual traffic. We systematically improve the performance of our MT systems quality on language pairs as measured by MT evaluation metrics including Bleu and Chrf to determine their impact on search precision metrics and extract signals that help to guide the  improvement strategies. Using this information we develop techniques to compare query translations for multiple language pairs and identify the most promising language pairs to invest and improve.
\end{abstract}

\section{Introduction}

Multilingual search capability is essential for modern e-commerce product discovery \citep{gartner2021,zhang-2022-improve}. Localization of e-commerce sites have led users to expect search engines to handle multilingual queries. Recent proposals of cross-lingual information retrieval handle multilingual queries, and language-agnostic cross-borders product indexing has gained traction with neural search engines \citep{hui-etal-2017-pacrr,mcdonald-etal-2018-deep,10.1145/3292500.3330759,lu-etal-2021-graph,li2021embedding}, but legacy e-commerce search indices are still built on monolingual product information and support for multilingual search is bridged using Query translation \citep{nie2010cross,10.1145/3308558.3313502,saleh-pecina-2020-document,bi2020constraint,jiang-etal-2020-cross, zhang-tan-2021-textual}.

Query translation allows users to look up information represented in documents written in a languages different from the language of the query. It takes as input the query typed in source or query language and returns a translated query to the search engine to retrieve documents in the target language. It follows that query translation plays a key role and its output significantly affects the retrieval results. 

Previous studies have demonstrated performance of CLIR (Cross-Lingual Information Retrieval) correlates highly with the quality of the Machine Translation (MT), and improving the quality of MT improves retrieval quality \citep{goldfarb2019artificial,brynjolfsson2019does}. However, these evaluations are done separately for each task. This leaves a large gap in understanding the impact of improving MT quality iteratively on 
 CLIR performance in a real time industrial setting. Since machine translation is used here as interim application, the objectives of the retrieval task may have varying levels of tolerance to the inherent translation quality. Information retrieval evaluation usually involves human-annotated relevance labels of search results candidates. In an industry setting, annotating a representative sample is a time consuming and expensive task, particularly during iterative improvement of MT for the search use case. Additionally, a general-purpose MT evaluation metric may not necessarily adapt to the query evaluation for downstream retrieval task.

To address these above concerns, we propose an MT evaluation framework to build an e-commerce specific CLIR test set. It exploits behavioural signals from search retrieval results to evaluate  MT quality for a given query. In order to identify the benefit of improving an MT system, we further investigate the sensitivity of retrieval quality  to the presence of different levels of MT quality as measured by Bleu, and Chrf using experimental datasets collected from actual traffic. Based on these experiments, we recommend the pairs that are worth continued investment in improving  MT systems for search.
Our main contributions are:

\begin{outline}
    \1 A \textbf{rank-based evaluation framework} to evaluate MT in CLIR through ranking-based search metrics using behavioral signals (from the store of the target language) as a proxy to relevance information without any human annotation; this framework can be used to create e-commerce CLIR test set at scale.
    \1 A \textbf{method to measure the MT launching impact} on the e-commerce CLIR ecosystems for a given language pair. This can be used to identify and prioritize the high impact language pairs for more investment in the MT improvement. 
    \1 A \textbf{method to measure the MT improvement impact} on the e-commerce CLIR ecosystems for a given language pair. It signals the strategy to be used for MT improvement, either a comprehensive strategy focusing on the overall query traffic or a specific one targeting a smaller percentage of query traffic or a combination of both strategies.
\end{outline}

This paper is organized as following:  we propose a rank-based evaluation framework in section \ref{section:ndcg}. We propose two MT impact rates, MT launching impact rate and MT improvement MT rate respectively in section \ref{sec:impact}. Section \ref{sec:experiment} is the experiment with 12 language pairs from 6 stores. Section \ref{sec:res} is the results and analysis. We defer related work to Section \ref{sec:related} where we compare it with our proposed work. We draw a conclusion in Section \ref{sec:conclusion}.

\section{Cross-Lingual Information Retrieval (CLIR) Evaluation Framework for E-commerce Product Search}
\label{section:ndcg}
Different from static test sets in academia, industrial search applications are dynamic as user queries and behavioral signals change with world trends. Moreover, product inventory is dynamic, changes often and quickly. 


A previous study \cite{sloto2018leveraging} proposes the traditional Normalized Discounted Cumulative Gain (nDCG) for CLIR using all search results from the reference translation as relevance ground truth to compute nDCG for MT translation (aka nDCG-MT). However, their approach imposes a strong assumption that the top-$k$ search results from reference translation are all relevant to the query and relevance is inversely scaled by the ranking of the results. 

Although behavioral signals from users’ clicks and purchases are useful proxy  \citep{10.1145/3209978.3209993} to expensive human relevance annotations, these are dynamic and change according to the product life cycle and seasonal business trends. These behavioral signals need to be updated at regular cadence to accurately represent relevance information needed to compute search metrics. 

We introduce a ranking-based evaluation framework through search ranking metrics using behavioral signals as a proxy to relevance information without any human annotation; To the best of our knowledge, there is no systematic study on cross-lingual information retrieval for e-commerce search that neither requires ground-truth click/purchase information nor human annotated relevance data.

\begin{figure}[h]
\includegraphics[width=8cm]{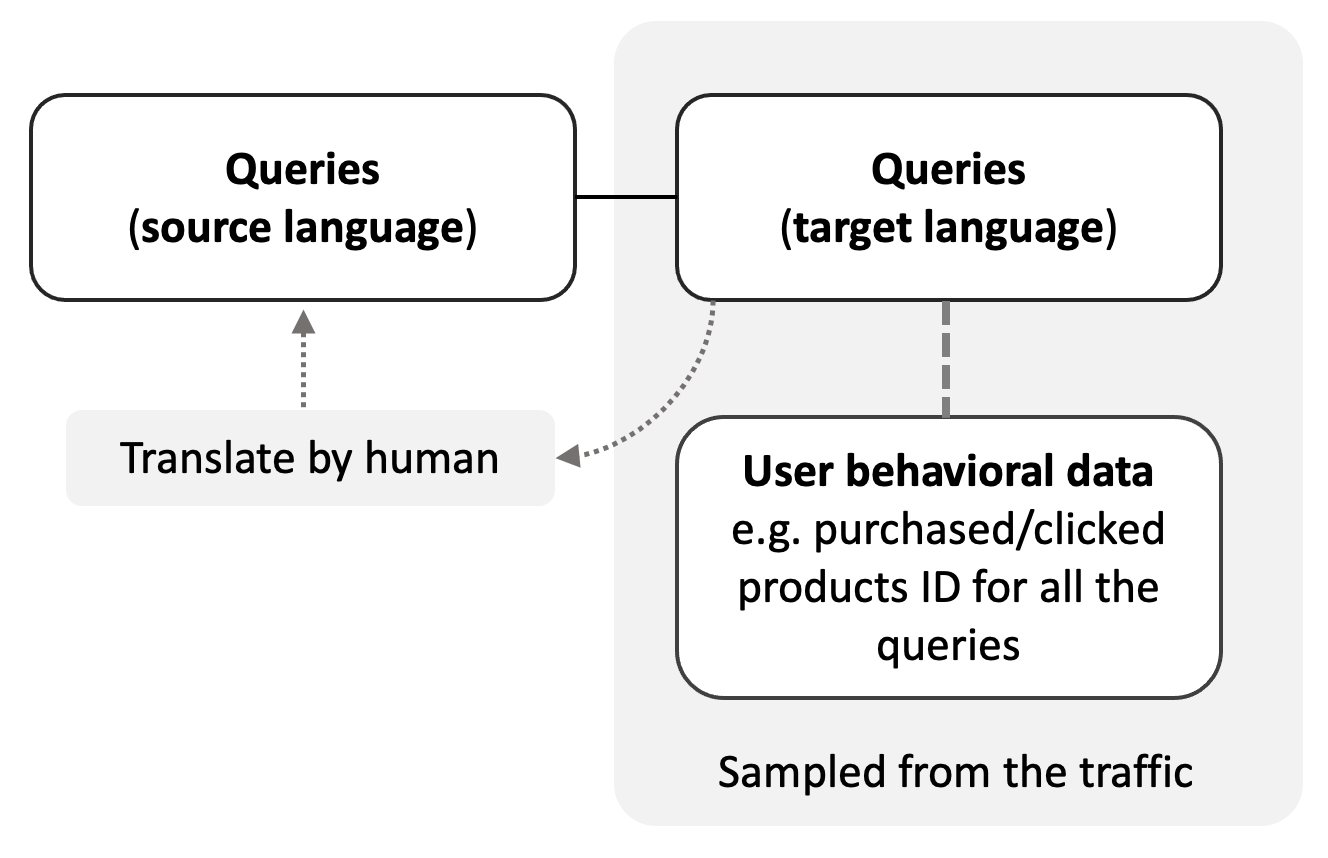}
\caption{Test set creation workflow }
\label{fig:testset}
\centering
\end{figure}

Figure \ref{fig:testset} illustrates the test sets creation workflow for MT evaluation in E-commerce CLIR:
\begin{outline}[enumerate]
    \1 Create a sample of query data from the historical search traffic in the target language (the language that the search index is built on). Empirically, we recommend to sample that queries from the top 30\%, bottom 30\% and the middle 40\% in frequency bins to better simulate the user traffic. \textit{We refer to these queries as $Q_{ref}$.}
   
        \1 To allow computation of traditional relevance metrics, aggregate the clicks and/or purchase product IDs associated with the queries, if they are available. We refer to the  products IDs associated with the query and their click/purchase frequency as $P_{id}$ and $P_{freq}$.
    
    \1 Create human reference translation of the search queries sample in the source language (the language that users will be searching in). \textit{We refer to these human translated queries as $Q_{src}$.}
\end{outline}
We propose the following evaluation framework with the test sets created above to evaluate machine translation in the context of CLIR for e-commerce queries. 

\begin{outline}[enumerate]
   
    
    

    

    \1 Translate the $Q_{src}$ with the MT model in consideration. \textit{We refer to these machine translated queries as $Q_{mt}$.}

    \1 Search for the  candidate products using the machine translated queries $Q_{mt}$; retrieving top-$k$ search result $R_{mt}$
    
    \1 Use $P_{id}$ and $P_{freq}$ (as ground truth) with $R_{mt}$ to compute traditional relevance based metrics such as nDCG.
    
\end{outline}










\section{MT impact in the search ecosystem}
\label{sec:impact}


\subsection{The range of MT impact on search}
\label{section: impact-range}
As mentioned, the downstream search pipeline consists of a large number of components, which altogether has different levels of tolerance for query translation quality. Hence, it is important to estimate the range of query translation impact on the search ecosystem in consideration. With the test sets from the creation workflow in Section \ref{section:ndcg}, we propose to measure the rank-based search metrics such as nDCG of source queries $Q_{src}$ as the lower bound of the MT impact, which serves a baseline for the impact of MT translated query on search, and measure the search metrics of human reference query translations $Q_{ref}$ as the upper bound. 

\subsection{MT launching impact measurement}
\label{section:impact-rate}
We expect that launching an MT system in a search ecosystem of different language pairs can have different levels of positive impact on the search result quality. Therefore, given a language pair (\textit{e.g. enus-jajp}), we propose the \textbf{MT launching impact rate} to quantify the MT impact on the search ecosystem (\textit{e.g. jp}). MT launching impact rate ($I_{MT}$) is defined as in Equation \ref{eq:impact-rate}. 

\begin{equation}
    I_{MT}=\frac{\Delta S}{\Delta T}
    \label{eq:impact-rate}
\end{equation}
\begin{equation}
    \Delta S = S_{adapt} - S_{source}
    \label{eq:q_s}
\end{equation}
\begin{equation}
    \Delta T = T_{adapt}-T_{source}
    \label{eq:q_t}
\end{equation}

\noindent where, $\Delta S$ is the search result improvement from source queries $S_{source}$ to the query translation from a fine-tuned MT $S_{adapt}$ (as Equation \ref{eq:q_s}), $S_{source}$ and $S_{adapt}$ can be common search metrics such as nDCG, $\Delta T$ is the respective  translation quality improvement from the source queries $T_{source}$ to the fine-tuned MT query translations $T_{adapt}$ (as equation \ref{eq:q_t}), $T_{source}$ and $T_{adapt}$ can be MT evaluation metrics such as Bleu or Chrf.

We propose the following three groups for language pairs based on their MT impact rate:
\begin{itemize}
    \item \noindent\textbf{High-impact language pairs}: Search ecosystems of high-impact language pairs are less tolerant to languages different from the search index language, and more sensitive to the query translation quality. Launching or improving an MT system of those language pairs in the respective search pipeline is more likely to improve the search results. 
    \item \noindent\textbf{Medium-impact language pairs}: Search ecosystems of medium-impact language pairs are somewhat sensitive to the query translation quality, though not as much as high-impact language pairs. 
    \item \noindent\textbf{Low-impact language pairs}: Search ecosystems of low-impact language pairs are more robust to different languages and translation quality, and the presence of an MT in the search pipeline has less or little impact on the search result improvement.
\end{itemize}




\subsection{MT improvement impact}



We experimented with two improvement strategies for MT in the e-commerce CLIR product search: one is \textbf{comprehensive improvement (CI)}, the other is \textbf{ specific improvement (SI)}.  CI usually focuses on the overall improvement in translation quality  and targets the entire query traffic. The CI strategies usually involve a change of model architecture or training techniques, etc; SI usually focuses on the improvement of the specific aspects of the query translation quality,  and targets a fraction of query traffic. The SI strategies are not necessarily language-agnostic, for example, it can be solving a smaller transliteration problem in a given language, or a brand term preservation improvement for a given language pair. 


We propose \textbf{The MT improvement impact rate}  to quantify the impact of MT comprehensive improvement  ($I_{improve}$)  on search improvement  as in Equation \ref{eq:adapt-impact}, which can provide signals to choose the right MT improvement strategy for a given language pair.
\begin{equation}
I_{improve}=\frac{\Delta S^\prime}{\Delta T^\prime}
    \label{eq:adapt-impact}
\end{equation}
\begin{equation}
\Delta S^\prime=S_{adapt} - S_{generic}
    \label{eq:q_s_p}
\end{equation}
\begin{equation}
\Delta T^\prime = T_{adapt}-T_{generic}
    \label{eq:q_t_p}
\end{equation}
where, $\Delta S^\prime$ is the search result improvement from generic MT query translations $S_{generic}$ to the fine-tuned MT query translations $S_{adapt}$ (as in Equation \ref{eq:q_s_p}), $S_{generic}$ and $S_{adapt}$ can be the common search metrics such as nDCG; $\Delta T^\prime$ is the respective translation quality improvement from generic MT query translations $T_{generic}$ to the fine-tuned MT query translations $T_{adapt}$ (as in Equation \ref{eq:q_t_p}), $T_{generic}$ and $T_{adapt}$ can be MT evaluation metrics such as Bleu or Chrf.

Language pairs with higher improvement rate signals both the CI and SI of MT are likely to have positive impact on search. Those with lower rate may benefit more from the focusing on SI for a targeted group of queries from the traffic.
\section{Experiment}
\label{sec:experiment}
\textbf{Language pairs and locales:}
We selected 12 language pairs from 6 stores for our experiments as seen in Table \ref{tab:lang-arcs}.
\begin{table}[h]\centering
\resizebox{0.8\columnwidth}{!}{%
\begin{tabular}{|ll|ll|}
\hline
\begin{tabular}[c]{@{}l@{}}\textbf{Lang pair}\end{tabular} & \textbf{Store} & \begin{tabular}[c]{@{}l@{}}\textbf{Lang pair}\end{tabular} & \textbf{Store} \\ \hline
esmx-enus & US    & ptpt-eses & Spain   \\ \hline
ptbr-enus & US    & frca-enca & Canada  \\ \hline
kokr-enus & US    & nlnl-dede & Germany \\ \hline
dede-enus & US    & trtr-dede & Germany \\ \hline
mlin-enin & India & engb-dede & Germany \\ \hline
knin-enin & India & enus-jajp & Japan   \\ \hline
\end{tabular}%
}
\caption{Selected 12 language pairs from 6 stores}
\label{tab:lang-arcs}
\end{table}

\noindent\textbf{Test data}: The test data is created as described in Section \ref{section:ndcg}. The test set comprises 4000 queries (as reference query translation) per store (e.g. enus), each query is translated into their respective language pairs (e.g. enus -> kokr, enus -> dede). We have also stored the purchased product IDs associated with the queries of the store (e.g. US). We use sampled purchased product ID associated with reference queries as relevant product,  and the logarithm of the frequencies of purchased product as the relevance score. 
\\
\noindent\textbf{Machine Translation (MT) models}: We trained two models per language pair: (i) a \textit{generic MT} system trained on general news and internal crawled data with (ii) a \textit{domain-specific MT} that is fined tuned on human translated search queries and synthetically generated query translations through back-translation. These in-house MT models are trained on proprietary data using vanilla transformer architecture \citep{vaswani2017attention} with Sockeye MT toolkit \citep{domhan-etal-2020-sockeye}.\footnote{For the purpose of this paper, we are less concerned with the accuracy of the MT models and more interested in the difference in the MT quality as per measured by traditional MT metrics and their evaluation based on our proposed framework. Thus the brevity in the model description.}\\
\noindent\textbf{Metric hyper-parameters}: We set $K$ to 16 for the top-$k$ search results, using the top-16 products in the search results to compute nDCG@16. 
\\
\noindent \textbf{MT metrics}: Tables \ref{tab:Bleu} and \ref{tab:chrf} in the appendix present the MT quality metrics Bleu\footnote{SacreBleu version 2.0.0 \citep{post-2018-call}} and Chrf; Table \ref{tab:ndcg} in the appendix presents search performance metric normalized nDCG@16.\footnote{Both the nDCG@16 and Chrf are scaled to 0-100 for the computation convenience}.

\noindent \textbf{MT launching and improvement impact rates}: With aforementioned metrics, the lower and higher bounds of nDCG@16 of MT impact are presented in Table \ref{tab:impact-range}. MT launching impact and improvement rates are computed using nDCG@16 with  and Chrf respectively, as in Table \ref{tab:all-impact} in the appendix.


\begin{table}[ht]\centering
\resizebox{\columnwidth}{!}{%
\begin{tabular}{|l|cc|cc|}
\hline
 &
  \multicolumn{2}{c|}{\textbf{\begin{tabular}[c]{@{}l@{}}MT launching\\impact\end{tabular}}} &
  \multicolumn{2}{l|}{\textbf{\begin{tabular}[c]{@{}l@{}}MT improvement\\impact\end{tabular}}} \\ \hline
\textbf{\begin{tabular}[c]{@{}l@{}}Language\\ pair\end{tabular}} &
  \multicolumn{1}{l|}{\textbf{\begin{tabular}[c]{@{}l@{}}$\Delta$nDCG/\\ $\Delta$Bleu\end{tabular}}} &
  \multicolumn{1}{l|}{\textbf{\begin{tabular}[c]{@{}l@{}}$\Delta$nDCG/\\ $\Delta$Chrf\end{tabular}}} &
  \multicolumn{1}{l|}{\textbf{\begin{tabular}[c]{@{}l@{}}$\Delta$nDCG/\\$\Delta$Bleu\end{tabular}}} &
  \multicolumn{1}{l|}{\textbf{\begin{tabular}[c]{@{}l@{}}$\Delta$nDCG/\\ $\Delta$Chrf\end{tabular}}} \\ \hline
ptpt-eses & \multicolumn{1}{c|}{0.11} & 0.15 & \multicolumn{1}{c|}{0.19} & 0.70 \\ \hline
enus-jajp & \multicolumn{1}{c|}{0.25} & 0.18 & \multicolumn{1}{c|}{0.78} & 1.09 \\ \hline
engb-dede & \multicolumn{1}{c|}{0.29} & 0.32 & \multicolumn{1}{c|}{0.09} & 0.13 \\ \hline
frca-enca & \multicolumn{1}{c|}{0.31} & 0.23 & \multicolumn{1}{c|}{0.35} & 0.60 \\ \hline
nlnl-dede & \multicolumn{1}{c|}{0.47} & 0.43 & \multicolumn{1}{c|}{0.32} & 0.69 \\ \hline
esmx-enus & \multicolumn{1}{c|}{0.50} & 0.34 & \multicolumn{1}{c|}{0.34} & 0.64 \\ \hline
ptbr-enus & \multicolumn{1}{c|}{0.62} & 0.56 & \multicolumn{1}{c|}{0.28} & 1.01 \\ \hline
dede-enus & \multicolumn{1}{c|}{0.62} & 0.66 & \multicolumn{1}{c|}{0.33} & 0.61 \\ \hline
knin-enin & \multicolumn{1}{c|}{0.72} & 0.59 & \multicolumn{1}{c|}{0.19} & 0.60 \\ \hline
trtr-dede & \multicolumn{1}{c|}{0.85} & 0.43 & \multicolumn{1}{c|}{0.24} & 0.43 \\ \hline
kokr-enus & \multicolumn{1}{c|}{0.98} & 0.49 & \multicolumn{1}{c|}{0.33} & 0.39 \\ \hline
mlin-enin & \multicolumn{1}{c|}{1.04} & 0.59 & \multicolumn{1}{c|}{0.74} & 0.72 \\ \hline
\end{tabular}%
}
\caption{MT launching impact and improvement impact rates}
\label{tab:all-impact}
\end{table}

\section{Results and Analysis }
\label{sec:res}
For the MT launching impact, we rank the language pairs in the descending order according to the MT launching impact rate as well as the impact range respectively, as in Table \ref{tab:impact-rank} in the appendix.  We observe Bleu and Chrf can give a similar ranking with small difference, so the following analysis is based on the MT launching impact from $\Delta$ nDCG/ $\Delta$Bleu for simplicity. For the MT improvement impact rate, we observe that Bleu makes value scale smaller than Chrf. We will use $\Delta$nDCG/ $\Delta$Bleu for the following analysis.

\begin{figure}[ht]\centering
    \includegraphics[scale=0.8]{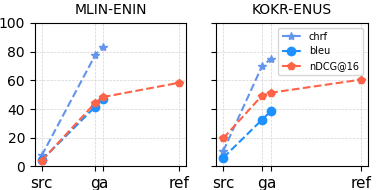}
    \caption{Language pairs where the MT have bigger impact on search pipeline}
    \label{fig:high-impact}
\end{figure}
\begin{figure}[ht]\centering
    \includegraphics[scale=0.8]{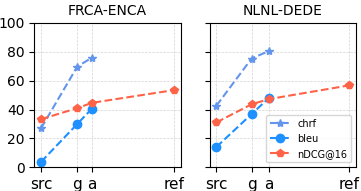}
    \caption{Language pairs where the MT have decent impact on search pipeline}
    \label{fig:ok-impact}
\end{figure}
\begin{figure}[ht]\centering
    \includegraphics[scale=0.8]{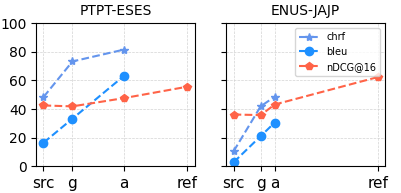}
    \caption{Language pairs where the MT have lower impact on search pipeline}
    \label{fig:low-impact}
\end{figure}

We observe that MT of language pairs such as \textit{mlin-enin, kokr-enus} have higher launching impact rate and should be labeled as the high-impact language pairs. For \textit{mlin-enin}, the MT launching impact rate is 1.04, which signals one point Bleu increase in translation quality can gain slightly more than one point of search improvement. Figure \ref{fig:high-impact} (In Figure \ref{fig:high-impact}, \ref{fig:ok-impact},  \ref{fig:low-impact} , ``src'' refers to  source query, ``g'' refers to generic mt, ``a'' refers to the adapted (fine-tuned) MT, ``ref'' refers to the human translation. The axis is scaled according to the Bleu score from 0-100.) illustrates the higher impact language pairs, the range of the MT impact is much bigger, search ecosystems are very responsive to the presence of MT system in the search pipeline, MT and search metrics have similar trending. \textit{mlin-enin} has a much higher improvement rate of 0.74, the ecosystem of the search of this language pair can potentially benefit from both comprehensive improvement (CI) and specific improvement (SI) in the MT. Meanwhile, \textit{kokr-enus} has a much lower improvement rate of 0.33, which signals this search is more likely to benefit from SI than CI.

\noindent Language pairs such as \textit{nlnl-dede, frca-enca} should be considered as the decent impact language pairs. As illustrated in Figure \ref{fig:ok-impact}, both  have smaller MT impact range and the launching impact rates are high but not quite as the high impact language pairs. As Bleu and Chrf increase from source query to generic MT to fine-tuned MT, nDCG@16 increases slower. Both language pairs have relative lower improvement impact rate which is around 0.3, that signals search of these two language pairs are more likely to benefit from SI than CI. 
Language pairs such as \textit{ptpt-eses, enus-jajp} should be labeled lower impact language pairs based on the lower launching impact rate. For \textit{ptpt-eses}, one point Bleu increase in translation quality can only achieve 0.11 point of search improvement. Both have smaller launching impact range, thus, search ecosystems are not very responsive to the MT quality improvement. As Bleu and Chrf increase from source query to generic MT to fine-tuned MT, nDCG@16 increases much slower and the trend line is almost flat as Figure \ref{fig:low-impact}. In principle, low-impact language arcs might not be prioritized for MT improvement. If there is a need to improve those MT for search, \textit{ptpt-eses}  has a much lower MT impact rate of 0.19, so search is likely to benefit from the SI for the MT, whereas \textit{enus-jajp} has much higher improvement rate of 0.78, the search may still benefit from CI as well as SI. 
Figure \ref{fig:plot1} and \ref{fig:plot2} in the appendix are the plots for all other language pairs.
\\ \\
\noindent\textbf{A/B testing:} We have also conducted parallel online A/B testing for the following language pairs: \textit{enus-jajp, ptpt-eses, frca-enca, mlin-enin, nlnl-dede, engb-dede}. For each language pair, we have deployed two fine-tuned MT systems and integrated them into the search pipeline for the designated store, and the MT system with the comprehensive improvement has higher off-line MT metrics (~+5 Bleu points on average) than the baseline model. The A/B testing lasted for 4 weeks on average for all the experiments. For the high impact language pairs, the improved MT systems  have seen large increases in  business metrics, such as, Order Product Sales (OPS), composite contribution profit (CCP), compared to the baseline model, and have much larger positive impact on the search result quality. 
For the low impact language pairs, we observe much smaller or even no impact at all. Overall, the A/B testing results are consistent with the MT launching impact rate results we have computed. Moreover, for ptpt-eses and nlnl-dede, we also conducted another round A/B testing with the same experiment setup except using MT with specific improvement to compare with the baseline models. Those two improved MT enhanced the terminology translation of 3-5\% of query traffic. The results are consistent with our hypothesis that the MT with SI improvement has much more impact than the MT with CI improvement.

\section{Related Work}
\label{sec:related}

Machine Translation is necessary to bridge the gap between query translation and cross-lingual information retrieval \cite{bi2020constraint}. Query translation a key component in large e-commerce stores, previous studies have demonstrated that better translation quality improves retrieval accuracy \citep{goldfarb2019artificial,brynjolfsson2019does}.

Queries are naturally short and search engines usually have preferred word choices and collocations based on users' query patterns \citep{lv2009adaptive,vechtomova2006study}. This complicates the evaluation of machine translation for cross-lingual information retrieval in the context of `fitting in well to the search index`. While machine translation evaluation is well-studied, translation evaluation in downstream task requires more attention especially in the e-commerce cross-lingual information retrieval.

Traditionally, information retrieval evaluation relies on behavioral signals as ground truth to measure relevance of search results; mean reciprocal ranking (MRR), mean average precision (MAP), normalized discounted cumulative gain (nDCG) \citep{jarvelin2002cumulated,10.1145/3209978.3209993, nigam2019semantic}.

Previous studies in cross-lingual information retrieval (CLIR) evaluation relies on pre-annotated datasets that are relatively small and specific to domains outside of e-commerce; for example, the CLEF eHealth test sets  \cite{saleh2018cuni,suominen2018overview,zhang2013comparative} and Wikipedia cross-lingual test set \cite{sas-etal-2020-wikibank}. Although \citet{sloto2018leveraging} proposed the nDCG-MT metric that leveraged on the reference translation to measure search results relevance, reliance on the ground truth data is still necessary. 
In pursuit of a more effective approach, we integrate CLIR and MT more closely and evaluate them in an end-to-end task. Our proposed method allows us to fully-automate the evaluation and study the impact of improving MT on CLIR by collecting organic queries in the target language of the e-commerce service and use reference results of these queries as a proxy to human annotation.
\section{Conclusion}
\label{sec:conclusion}
In this paper, we propose an evaluation framework for MT in the E-commerce multilingual product search through ranking-based search metrics using behavioral signals as proxy relevance information without any human notation, which makes it practical to iteratively improve MT models for the search use case and evaluate them frequently off-line. This framework can also be used to create cross-lingual information retrieval (CLIR) test sets for e-commerce at scale. We also propose a method to measure off-line the MT launching impact and and improvement impact rate on search. The former can identify the the high-impact language pairs can be prioritized with more investment in the MT improvement. These experiments can help select the most promising improvement strategy either comprehensive or specific improvement or combination of both to bring a larger impact on the search performance of a given language pair. We have experimented with the proposed evaluation framework and MT impact measuring method on 12 language pairs from 6 stores, and identified the high language pairs of different impact on search and assigned potential improvement strategies. The results are consistent with on-line A/B testing.

\bibliographystyle{acl_natbib}
\bibliography{anthology,others,acl2021,reference}
\newpage
\appendix
\section{Appendix}

\label{sec:appendix}

\begin{table}[ht] \centering
\resizebox{0.8\columnwidth}{!}{%
\begin{tabular}{|llll|}
\hline
\multicolumn{4}{|c|}{\textbf{sacreBleu}}                                                                   \\ \hline
\multicolumn{1}{|l|}{\textbf{\begin{tabular}[c]{@{}l@{}}Language \\ pair\end{tabular}}} &
  \multicolumn{1}{l|}{\textbf{source}} &
  \multicolumn{1}{l|}{\textbf{\begin{tabular}[c]{@{}l@{}}generic\\ MT\end{tabular}}} &
  \textbf{\begin{tabular}[c]{@{}l@{}}adapted \\ MT\end{tabular}} \\ \hline
\multicolumn{1}{|l|}{trtr-dede} & \multicolumn{1}{l|}{6.4}   & \multicolumn{1}{l|}{23.4}  & 28.8  \\ \hline
\multicolumn{1}{|l|}{enus-jajp} & \multicolumn{1}{l|}{2.8}   & \multicolumn{1}{l|}{21.1}  & 30.6  \\ \hline
\multicolumn{1}{|l|}{esmx-enus} & \multicolumn{1}{l|}{2.6}   & \multicolumn{1}{l|}{26.6}  & 33.3  \\ \hline
\multicolumn{1}{|l|}{kokr-enus} & \multicolumn{1}{l|}{6.02}  & \multicolumn{1}{l|}{32.53} & 38.39 \\ \hline
\multicolumn{1}{|l|}{frca-enca} & \multicolumn{1}{l|}{3.77}  & \multicolumn{1}{l|}{30.01} & 40.46 \\ \hline
\multicolumn{1}{|l|}{ptbr-enus} & \multicolumn{1}{l|}{3.7}   & \multicolumn{1}{l|}{26.8}  & 41.91 \\ \hline
\multicolumn{1}{|l|}{mlin-enin} & \multicolumn{1}{l|}{4.41}  & \multicolumn{1}{l|}{41.7}  & 47.02 \\ \hline
\multicolumn{1}{|l|}{nlnl-dede} & \multicolumn{1}{l|}{14.09} & \multicolumn{1}{l|}{36.87} & 48.11 \\ \hline
\multicolumn{1}{|l|}{dede-enus} & \multicolumn{1}{l|}{6.88}  & \multicolumn{1}{l|}{46.74} & 60.93 \\ \hline
\multicolumn{1}{|l|}{ptpt-eses} & \multicolumn{1}{l|}{16.49} & \multicolumn{1}{l|}{33.28} & 63.08 \\ \hline
\multicolumn{1}{|l|}{engb-dede} & \multicolumn{1}{l|}{10.1}  & \multicolumn{1}{l|}{45.61} & 63.08 \\ \hline
\multicolumn{1}{|l|}{knin-enin} & \multicolumn{1}{l|}{2.77}  & \multicolumn{1}{l|}{52.02} & 71.27 \\ \hline
\end{tabular}%
}
\caption{MT metric - Bleu for source queries and query MT translations }
\label{tab:Bleu}
\end{table}


\begin{table}[ht]\centering
\resizebox{0.8\columnwidth}{!}{%
\begin{tabular}{|llll|}
\hline
\multicolumn{4}{|c|}{\textbf{Chrf}}                                                               \\ \hline
\multicolumn{1}{|l|}{\textbf{\begin{tabular}[c]{@{}l@{}}language\\ pair\end{tabular}}} &
  \multicolumn{1}{l|}{\textbf{source}} &
  \multicolumn{1}{l|}{\textbf{\begin{tabular}[c]{@{}l@{}}generic \\ mt\end{tabular}}} &
  \textbf{\begin{tabular}[c]{@{}l@{}}adapted \\ mt\end{tabular}} \\ \hline
\multicolumn{1}{|l|}{dede-enus} & \multicolumn{1}{l|}{30.49} & \multicolumn{1}{l|}{73.82} & 81.36 \\ \hline
\multicolumn{1}{|l|}{engb-dede} & \multicolumn{1}{l|}{33.08} & \multicolumn{1}{l|}{69.68} & 80.99 \\ \hline
\multicolumn{1}{|l|}{enus-jajp} & \multicolumn{1}{l|}{10.49} & \multicolumn{1}{l|}{41.91} & 48.67 \\ \hline
\multicolumn{1}{|l|}{esmx-enus} & \multicolumn{1}{l|}{24.92} & \multicolumn{1}{l|}{65.62} & 69.19 \\ \hline
\multicolumn{1}{|l|}{frca-enca} & \multicolumn{1}{l|}{27.04} & \multicolumn{1}{l|}{69.72} & 75.85 \\ \hline
\multicolumn{1}{|l|}{kokr-enus} & \multicolumn{1}{l|}{10.7}  & \multicolumn{1}{l|}{69.79} & 74.75 \\ \hline
\multicolumn{1}{|l|}{mlin-enin} & \multicolumn{1}{l|}{7.64}  & \multicolumn{1}{l|}{77.7}  & 83.19 \\ \hline
\multicolumn{1}{|l|}{nlnl-dede} & \multicolumn{1}{l|}{42.63} & \multicolumn{1}{l|}{75.34} & 80.58 \\ \hline
\multicolumn{1}{|l|}{ptbr-enus} & \multicolumn{1}{l|}{25.66} & \multicolumn{1}{l|}{64.2}  & 68.33 \\ \hline
\multicolumn{1}{|l|}{ptpt-eses} & \multicolumn{1}{l|}{48.37} & \multicolumn{1}{l|}{73.26} & 81.52 \\ \hline
\multicolumn{1}{|l|}{trtr-dede} & \multicolumn{1}{l|}{23.08} & \multicolumn{1}{l|}{64.27} & 67.3  \\ \hline
\multicolumn{1}{|l|}{knin-enin} & \multicolumn{1}{l|}{5.29}  & \multicolumn{1}{l|}{82.67} & 88.62 \\ \hline
\end{tabular}%
}
\caption{MT metric -Chrf for source queries and query MT translations}
\label{tab:chrf}
\end{table}

\begin{table}[t] \centering
\resizebox{0.9\columnwidth}{!}{%
\begin{tabular}{|lllll|}
\hline
\multicolumn{5}{|c|}{\textbf{nDCG@16}}                                                                                            \\ \hline
\multicolumn{1}{|l|}{\textbf{\begin{tabular}[c]{@{}l@{}}Language \\ pair\end{tabular}}} &
  \multicolumn{1}{l|}{\textbf{source}} &
  \multicolumn{1}{l|}{\textbf{\begin{tabular}[c]{@{}l@{}}generic \\ MT\end{tabular}}} &
  \multicolumn{1}{l|}{\textbf{\begin{tabular}[c]{@{}l@{}}adapted \\ MT\end{tabular}}} &
  \textbf{ref} \\ \hline
\multicolumn{1}{|l|}{enus-jajp} & \multicolumn{1}{l|}{36.2}  & \multicolumn{1}{l|}{35.80} & \multicolumn{1}{l|}{43.19} & 62.30 \\ \hline
\multicolumn{1}{|l|}{frca-enca} & \multicolumn{1}{l|}{33.34} & \multicolumn{1}{l|}{40.98} & \multicolumn{1}{l|}{44.64} & 53.47 \\ \hline
\multicolumn{1}{|l|}{trtr-dede} & \multicolumn{1}{l|}{26.8}  & \multicolumn{1}{l|}{44.60} & \multicolumn{1}{l|}{45.90} & 63.90 \\ \hline
\multicolumn{1}{|l|}{nlnl-dede} & \multicolumn{1}{l|}{31.11} & \multicolumn{1}{l|}{43.67} & \multicolumn{1}{l|}{47.26} & 56.76 \\ \hline
\multicolumn{1}{|l|}{ptpt-eses} & \multicolumn{1}{l|}{42.53} & \multicolumn{1}{l|}{41.89} & \multicolumn{1}{l|}{47.64} & 55.65 \\ \hline
\multicolumn{1}{|l|}{mlin-enin} & \multicolumn{1}{l|}{4.00}  & \multicolumn{1}{l|}{44.38} & \multicolumn{1}{l|}{48.34} & 58.28 \\ \hline
\multicolumn{1}{|l|}{ptbr-enus} & \multicolumn{1}{l|}{27.2}  & \multicolumn{1}{l|}{46.71} & \multicolumn{1}{l|}{50.89} & 60.28 \\ \hline
\multicolumn{1}{|l|}{kokr-enus} & \multicolumn{1}{l|}{19.59} & \multicolumn{1}{l|}{49.38} & \multicolumn{1}{l|}{51.29} & 60.42 \\ \hline
\multicolumn{1}{|l|}{dede-enus} & \multicolumn{1}{l|}{17.78} & \multicolumn{1}{l|}{46.91} & \multicolumn{1}{l|}{51.54} & 60.27 \\ \hline
\multicolumn{1}{|l|}{knin-enin} & \multicolumn{1}{l|}{2.90}  & \multicolumn{1}{l|}{48.7}  & \multicolumn{1}{l|}{52.27} & 58.28 \\ \hline
\multicolumn{1}{|l|}{esmx-enus} & \multicolumn{1}{l|}{37.7}  & \multicolumn{1}{l|}{50.6}  & \multicolumn{1}{l|}{52.90} & 69.40 \\ \hline
\multicolumn{1}{|l|}{engb-dede} & \multicolumn{1}{l|}{38.54} & \multicolumn{1}{l|}{52.38} & \multicolumn{1}{l|}{53.88} & 61.91 \\ \hline
\end{tabular}%
}
\caption{search metric (nDCG@16) of source queries and query MT and reference translations}
\label{tab:ndcg}
\end{table}

\begin{table}[h] \centering
\resizebox{0.8\columnwidth}{!}{%
\begin{tabular}{|l|l|l|l|}
\hline
\textbf{\begin{tabular}[c]{@{}l@{}}Language\\ pair\end{tabular}} &
  \textbf{\begin{tabular}[c]{@{}l@{}}lower\\ bound\end{tabular}} &
  \textbf{\begin{tabular}[c]{@{}l@{}}upper\\ bound\end{tabular}} &
  \textbf{\begin{tabular}[c]{@{}l@{}}impact\\ range\end{tabular}} \\ \hline
ptpt-eses & 42.53 & 55.65 & 13.12 \\ \hline
frca-enca & 33.34 & 53.47 & 20.13 \\ \hline
engb-dede & 38.54 & 61.91 & 23.37 \\ \hline
nlnl-dede & 31.11 & 56.76 & 25.65 \\ \hline
enus-jajp & 36.20 & 62.30 & 26.10 \\ \hline
esmx-enus & 37.70 & 69.40 & 31.70 \\ \hline
ptbr-enus & 27.20 & 60.28 & 33.08 \\ \hline
trtr-dede & 26.80 & 63.90 & 37.10 \\ \hline
kokr-enus & 19.59 & 60.42 & 40.83 \\ \hline
dede-enus & 17.78 & 60.27 & 42.49 \\ \hline
mlin-enin & 4.00  & 58.28 & 54.28 \\ \hline
knin-enin & 2.90  & 58.28 & 55.38 \\ \hline
\end{tabular}%
}
\caption{The MT impact range (nDCG@16) }
\label{tab:impact-range}
\end{table}

\begin{table}[ht]\centering

\resizebox{0.8\columnwidth}{!}{%
\begin{tabular}{|l|l|ll|}
\hline
  &           & \multicolumn{2}{c|}{\textbf{MT launching impact}} \\ \hline
\textbf{Rank} &
  \multicolumn{1}{c|}{\textbf{\begin{tabular}[c]{@{}c@{}}impact\\ range\end{tabular}}} &
  \multicolumn{1}{l|}{\textbf{\begin{tabular}[c]{@{}l@{}}$\Delta$nDCG/\\ $\Delta$Bleu\end{tabular}}} &
  \textbf{\begin{tabular}[c]{@{}l@{}}$\Delta$nDCG/\\ $\Delta$Chrf\end{tabular}} \\ \hline
1  & knin-enin & \multicolumn{1}{l|}{mlin-enin}     & dede-enus    \\ \hline
2  & mlin-enin & \multicolumn{1}{l|}{kokr-enus}     & knin-enin    \\ \hline
3  & dede-enus & \multicolumn{1}{l|}{trtr-dede}     & mlin-enin    \\ \hline
4  & kokr-enus & \multicolumn{1}{l|}{knin-enin}     & ptbr-enus    \\ \hline
5  & trtr-dede & \multicolumn{1}{l|}{dede-enus}     & kokr-enus    \\ \hline
6  & ptbr-enus & \multicolumn{1}{l|}{ptbr-enus}     & trtr-dede    \\ \hline
7  & esmx-enus & \multicolumn{1}{l|}{esmx-enus}     & nlnl-dede    \\ \hline
8  & enus-jajp & \multicolumn{1}{l|}{nlnl-dede}     & esmx-enus    \\ \hline
9  & nlnl-dede & \multicolumn{1}{l|}{frca-enca}     & engb-dede    \\ \hline
10 & engb-dede & \multicolumn{1}{l|}{engb-dede}     & frca-enca    \\ \hline
11 & frca-enca & \multicolumn{1}{l|}{enus-jajp}     & enus-jajp    \\ \hline
12 & ptpt-eses & \multicolumn{1}{l|}{ptpt-eses}     & ptpt-eses    \\ \hline
\end{tabular}%
}
\caption{Language pair ranking based on the MT launching impact}
\label{tab:impact-rank}
\end{table}

\newpage
\begin{figure*}[h]
    \centering
    \includegraphics[scale=0.8]{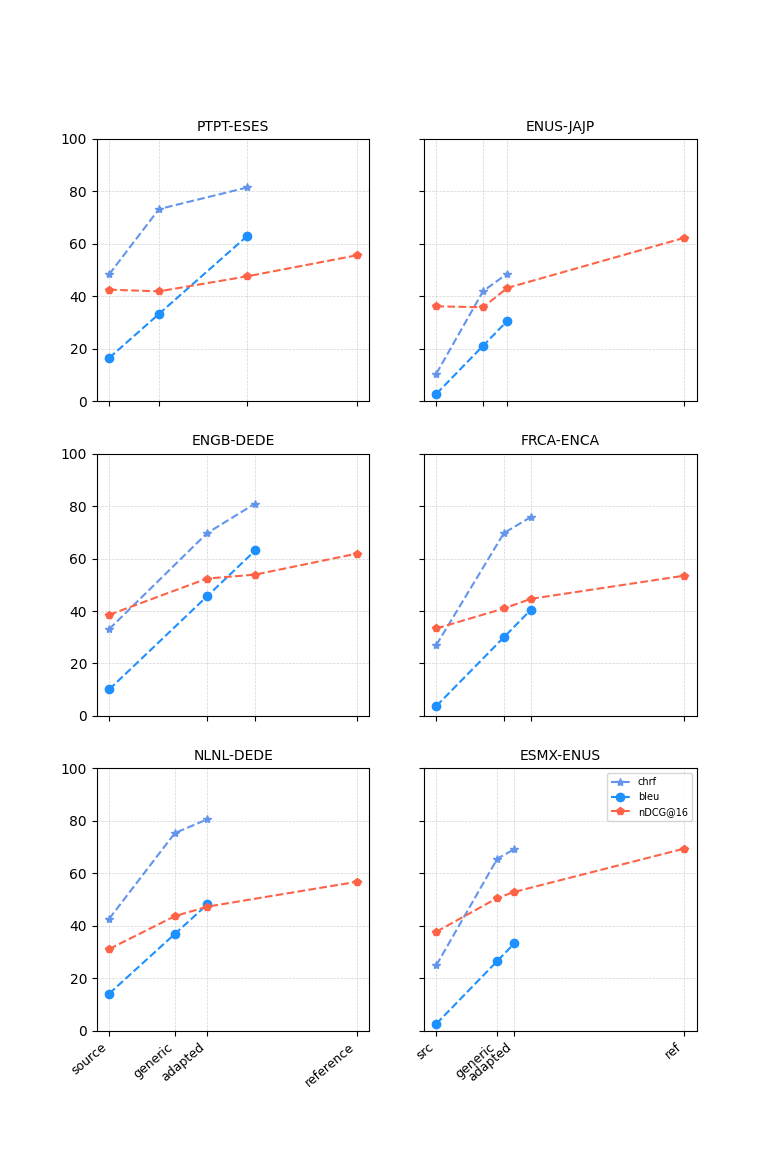}
    \caption{MT quality metrics and search metrics.png}
    \label{fig:plot1}
\end{figure*}

\begin{figure*}[h]
    \centering
    \includegraphics[scale=0.8]{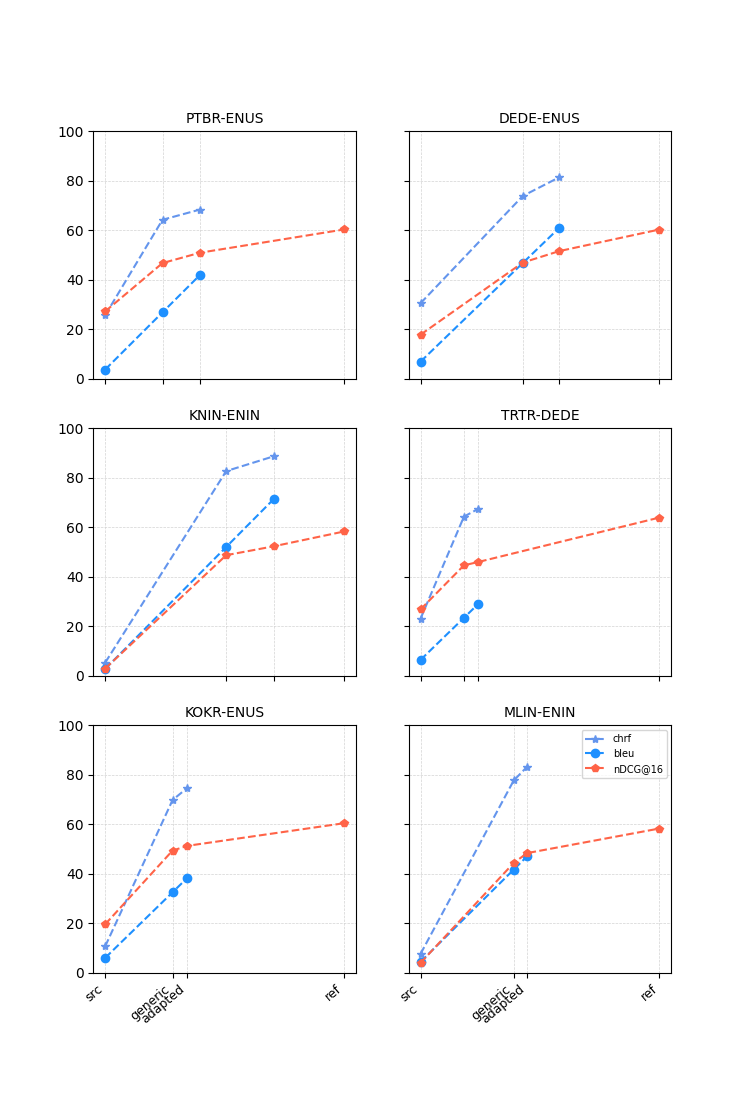}
    \caption{MT quality metrics and search metrics}
    \label{fig:plot2}
\end{figure*}

\end{document}